\documentclass[letterpaper, 10 pt, conference]{ieeeconf}  

\IEEEoverridecommandlockouts                              

\usepackage{graphics} 
\usepackage{epsfig} 
\usepackage{times} 
\usepackage{amsmath} 
\usepackage{amssymb}  
\usepackage{orcidlink} 
\usepackage{cite} %
\usepackage{mathtools} 
\usepackage{bm}
\usepackage{comment}
\usepackage{xcolor}
\usepackage{algorithm}
\usepackage{algpseudocode} 
\usepackage{svg}
\usepackage{placeins}
\usepackage{hyperref}  
\svgsetup{
  inkscape="C:/Program Files/Inkscape/bin/inkscape.exe",
}
\svgpath{{Figures/}}

\title{\LARGE \bf
MPC-based Coarse-to-Fine Motion Planning for Robotic Object Transportation in Cluttered Environments}

\author{Chen Cai$^{\orcidlink{0009-0002-3474-6671}}$$^{1}$, Ernesto Dickel Saraiva$^{\orcidlink{0009-0003-6680-2559}}$$^{1}$, Ya-Jun Pan$^{\orcidlink{0000-0002-8700-0956}}$$^{2}$, and Steven Liu$^{\orcidlink{0000-0002-3407-2745}}$$^{1}$
\thanks{$^{1}$Chen Cai, Ernesto Dickel Saraiva and Steven Liu are with
	the Department of Electrical and Computer Engineering, University of Kaiserslautern Landau,
	Kaiserslautern, 67663, Germany. 
	email:{\tt\small \{chen.cai, saraiva, steven.liu\}@rptu.de}}%
\thanks{$^{2}$Ya-Jun Pan is with the Department of Mechanical Engineering, Dalhousie
	University, Halifax, NS B3H 4R2, Canada
    (email: {\tt\small Yajun.Pan@dal.ca})}%
}

\begin{document}

\maketitle
\thispagestyle{empty}
\textit{This work has been submitted to the IEEE for possible publication. Copyright may be transferred without notice, after which this version may no longer be accessible.}
\pagestyle{empty}

\begin{abstract}This letter presents a novel coarse-to-fine motion planning framework for robotic manipulation in cluttered, unmodeled environments. The system integrates a dual-camera perception setup with a B-spline-based model predictive control (MPC) scheme. Initially, the planner generates feasible global trajectories from partial and uncertain observations. As new visual data are incrementally fused, both the environment model and motion planning are progressively refined. A vision-based cost function promotes target-driven exploration, while a refined kernel-perceptron collision detector enables efficient constraint updates for real-time planning. The framework accommodates closed-chain kinematics and supports dynamic replanning. Experiments on a multi-arm platform validate its robustness and adaptability under uncertainties and clutter.
\end{abstract}

\vspace{-1mm}
\section{INTRODUCTION}

Autonomous robots play a pivotal role in the transition towards Industry 4.0, driven by labor shortages exacerbated by the global pandemic and aging populations. Collaborative robots, designed to work with both humans and machines, are key to tackling these challenges. Compared to single manipulators, networked collaborative robots can handle larger and heavier objects, greatly improving operational efficiency.

However, the integration of multiple collaborative robots introduces heightened complexity in system configuration. which not only expands the dimensionality of the configuration space (C-space) but also complicates motion planning. Particularly in cluttered environments, cooperative object transportation can involve closed-chain constraints that restrict the available C-space. These constraints necessitate the careful design of sophisticated motion planners to manage the intricate dynamics of closed-chain systems effectively.

In this work, we propose a model predictive control~(MPC)-based motion planning framework for cooperative multi-arm object transportation in cluttered and unmodeled environments. To handle incomplete environment and target observations, the planner follows a coarse-to-fine strategy, gradually refining trajectories based on continuous updates from a dual-camera perception system. This integration enables robust, whole-body collision avoidance and target-aware motion generation under uncertainty, and is experimentally validated on a real-world multi-arm robotic setup operating in cluttered environments.

\section{Related Work}
In quest to develop autonomous robots, motion planning remains a pivotal challenge, particularly for high-dimensional redundant manipulators involved in whole-body motion planning. The computational complexity of navigating the C-space for such systems is considered NP-hard due to its high degrees of freedom (DoF), which greatly complicates the topology. Moreover, the presence of kinematic and dynamic constraints segments the state space into feasible and infeasible regions, further complicating planning efforts.

Significant research efforts have focused on sampling-based motion planning (SBMP), which sample the C-space to identify viable paths. The core advantage of SBMP, such as Probabilistic Roadmaps (PRM) and Rapidly-exploring Random Trees (RRT), lies in their probabilistic completeness: given sufficient time and resolution, they can theoretically find a solution if one exists~\cite{orthey2023sampling}. However, these methods are challenged by the curse of dimensionality, resulting in high computational demands and slow convergence.

To enhance the functionality of SBMP in heterogeneous environments, variations have been developed to optimize the search process. Techniques such as biased sampling \cite{denny2020dynamic} and informed-RRT* \cite{gammell2014informed} guide the exploration more effectively towards promising regions in C-space, utilizing initial feasible paths as heuristics to improve convergence rates and the quality of the solution paths. To address the challenges imposed by constraints, the Constrained Bidirectional RRT (CBi-RRT) technique extends traditional RRT by using projections to explore the constraint manifolds in C-space and to bridge between them \cite{berenson2009manipulation}. Despite these advancements, SBMP methods primarily address static or quasi-static environments and struggle with the dynamic constraints essential for kinodynamic planning, where velocity, acceleration, and torque constraints critically influence feasibility.

Another category within the spectrum of motion planning methods includes reactive approaches such as Artificial Potential Fields (APF) \cite{khatib1986real} and dynamical system-based methods \cite{khansari2012dynamical}. These are engineered to provide immediate obstacle avoidance by dynamically modulating trajectories around obstacles. However, while these methods excel at local path adjustments, they are inherently limited by their lack of global planning capabilities. Reactive methods often leads to suboptimal paths due to its focus on immediate surroundings rather than the entire route from start to finish. Furthermore, approaches based on APF, relying on artificial attractive and repulsive potentials, are prone to the local minima. Due to the excessive focus on nearby obstacles \cite{alhaddad2024neural}, robots using APF often fail to progress when repulsive forces outweigh the attractive force\cite{tang2024obstacle}. Dynamical system-based methods \cite{koptev2024reactive}, though theoretically ensuring convergence, can also encounter deadlock situations in cluttered or densely populated scenarios, requiring specialized modulation strategies \cite{conzelmann2022dynamical}. Beyond reactive methods, optimization-based planners such as GPMP2~\cite{pan2025survey} use Gaussian processes to generate smooth, probabilistically consistent trajectories, and have been applied effectively in collaborative robot settings. 

For motion planning in unmodeled and cluttered environments--such as reaching into a densely packed shelf where obstacles may not be fully visible or mapped--neither the complete workspace geometry nor the exact target pose is initially available. These conditions impose two fundamental challenges: ensuring generally safe and feasible motion despite incomplete knowledge, and the need to efficiently adapt as new information becomes available through perception.

Sampling-based planners, while effective for global path discovery, are often inefficient in dynamic or partially known environments due to dense sampling. Reactive methods like potential fields often suffer from local minima, especially in cluttered scenes where competing repulsive forces obscure target path. They are also sensitive to noisy or partial observations. Moreover, fully committing to fine-scale planning from the outset imposes heavy computational and perceptual costs, while purely coarse planning lacks the precision needed for safe manipulation and obstacle avoidance in tight, partially known spaces. This calls for a planning strategy that incrementally refines resolution with ongoing perception updates, balancing initial feasibility with progressive precision.

To address these challenges, we propose a coarse-to-fine motion planning strategy via a novel shrinking-horizon MPC framework based on B-spline collocation. We choose an MPC-based approach due to its capability for real-time, closed-loop updates and direct integration of constraints. At the initial stage, where environment knowledge and target poses are incomplete, the planner generates globally feasible trajectories from rough initial estimates. In parallel, a dual-camera perception system—combining stationary and eye-in-hand cameras—incrementally refines the environment model through data fusion and structured point cloud processing.

As perception updates accumulate, the MPC progressively refines its planned motion, transitioning from coarse to increasingly precise end-to-end trajectories. This framework supports dynamic replanning and integrates newly perceived information into the motion constraints in real time. An enhanced kernel-based collision detection module, adapted from~\cite{zhi2022diffco}, ensures efficient and accurate constraint evaluation in high-dimensional, closed-chain configurations, enabling safe and tractable navigation in cluttered, unstructured scenarios. The contributions are considered as follows:

\begin{itemize}
	\item Dual-Camera Perception with Incremental Model Refinement:
	Structured point cloud processing leveraging robot geometry for data filtering, environment segmentation into convex hulls, and occlusion polytopes generation, maintaining detailed workspace representations.
	\item Vision-Guided Shrinking-Horizon MPC: A novel MPC integrating vision-based costs and B-spline collocation, explicitly designed for uncertain environments with partially observed targets. In line with the ``coarse-to-fine'' strategy, it progressively refining coarse trajectories toward precise localization and constraint satisfaction.
	\item Efficient Collision Detection in High-DoF C-Space: Significant improvements over kernel-based collision detection \cite{zhi2022diffco} via refined active learning and kernel redesign, enabling faster, more accurate checking for high-dimensional, closed-chain robotic systems.
	\item Visibility-Aware Active Exploration:
	Differentiable visibility scoring using occlusion polytopes, optimizing eye-in-hand camera positions to systematically resolve uncertainties and discover partially visible targets.
	\item Implementation and Evaluation on a Multi-Arm Setup: Evaluation on a multi-arm setup with two Franka Emika (FE) cobots demonstrates real-time applicability, efficiency, and adaptability in cluttered environments. The system also responds to gradual scene changes, as shown in a human-intrusion scenario triggering safety-aware MPC replanning (Section~\ref{Scenario II}).
\end{itemize}

\section{Problem Formulation}
\begin{figure}[htbp]
	\centering
	\vspace{-3mm}
	\includegraphics[width = 0.98\linewidth]{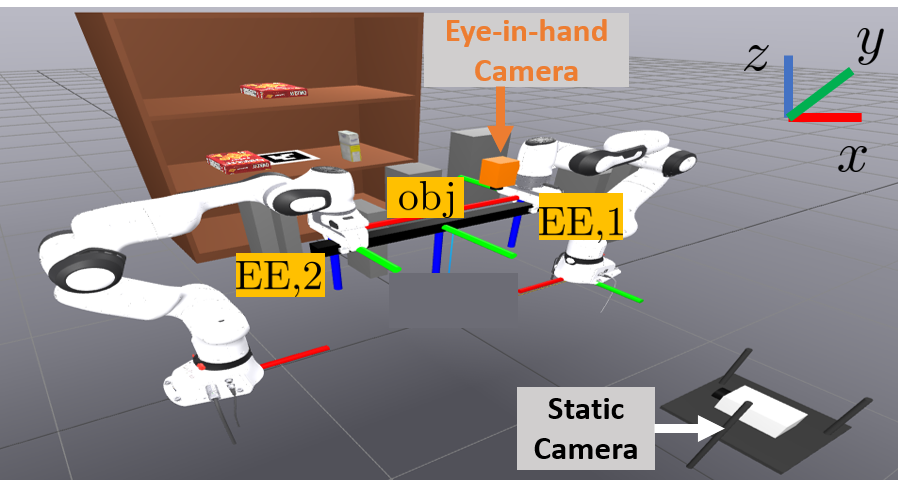}
	\vspace{-3mm}
	\caption{Cooperative transport in unmodeled clutter. The AprilTag-marked target on the shelf is initially occluded even for the dual-camera system.}
	\label{fig:Overview}
	\vspace{-2mm}
\end{figure}

Fig.~\ref{fig:Overview} illustrates the scenario addressed in this study, featuring a multi-arm robotic system cooperatively transports a rigid object in an unmodeled, cluttered environment. Unlike conventional setups with predefined targets, the system must handle targets that are partially observable, invisible, or dynamically changing. Although rough initial guesses from partial features or edges can be autonomously obtained, accurately locating the target and understanding the unmodeled environment require further processing and exploration.

Operating effectively in such an unstructured and dynamically changing environment presents significant challenges, particularly regarding reliable perception and motion planning. Addressing these challenges requires robust real-time perception and adaptive planning. To this end, we adopt a dual-camera setup consisting of a stationary camera and an eye-in-hand camera. This combination mitigates the limitations associated with a single fixed viewpoint, enhancing visibility through multi-view perception. Specifically, the stationary camera provides a stable, broad overview of the environment, while the eye-in-hand camera dynamically adjusts its viewpoint during robotic movements, progressively clarifying initially partially observed targets and environmental features. Moreover, this multi-view fusion enhances perception reliability and responsiveness, especially in the presence of dynamic obstacles or human workers.


\subsection{Preliminary}
\label{SecPreliminary}
   For the $i$-th redundant manipulator arm agent of $n$ DoF, its joint space dynamics model is described by \cite{cai2021task}:
\begin{equation}
	\bm{M}_i(\bm{q}_i)\ddot{\bm{q}}_i + \bm{C}_i(\bm{q}_i, \dot{\bm{q}}_i)\dot{\bm{q}_i} + \bm{\tau}_{\text{g}, i}(\bm{q}_i) = \bm{\tau}_i + \bm{\tau}_{\text{ext}, i} + \bm{\tau}_{\text{f}, i},
	\label{eqa: robot_system_kinematics}
\end{equation}
with the generalized joint coordinates $\bm{q}_i \in \mathbb{R}^{\raisebox{0.4ex}{$\scriptstyle n$}}$, generalized actuator torques $\bm{\tau}_i \in \mathbb{R}^{\raisebox{0.4ex}{$\scriptstyle n$}}$ and external torques $\bm{\tau}_{\text{ext}, i} \in \mathbb{R}^{\raisebox{0.4ex}{$\scriptstyle n$}}$. The inertia matrix $\bm{M}_i$, the Coriolis terms $\bm{C}_i$, the gravity torque vector $\bm{\tau}_{\text{g}, i}$ and the friction torque $\bm{\tau}_{\text{f}, i}$ are all configuration-dependent. 
The system output function defines the pose (position $\mathbf{x}_{p,i}$ and orientation $\mathbf{x}_{o,i}$ in quaternion) of $i$-th robot's end effector in Cartesian coordinates as follows:
\begin{equation}
	\mathbf{x}_i = \begin{bmatrix} \mathbf{x}_{p,i} \\ \mathbf{x}_{o,i} \end{bmatrix} = \begin{bmatrix} \bm{h}_{p,i}(\bm{q}_i) \\ \bm{h}_{o,i}(\bm{q}_i) \end{bmatrix} = \bm{h}_i(\bm{q}_i).
	\label{eqa: robot_system_output}
\end{equation}
Here, $\bm{h}_i$ represent the forward kinematics mapping. In non-singular configurations, the derivative of (\ref{eqa: robot_system_output}) yields the differential kinematics, defined by the Jacobian $\mathbf{J}_i(\bm{q}) = \frac{\partial \bm{h}_i}{\partial \bm{q}}$.

We assume rigid grasping, meaning no relative motion between each robot's grippers and the object, forming a closed-chain configuration in the multi-robot system. Given predefined grasping points, the relative pose between the object and the end-effector of the $i$-th robot is represented by the transformation matrix \( \prescript{\text{EE}, i}{}{T}_\text{obj} \). As depicted in Fig.~\ref{fig:Overview}, the object frame is denoted by the subscript ``obj'', the end effector by ``EE'', and the base frame by $i$. The object pose is determined using the direct kinematics of each manipulator:
\begin{equation}
	\prescript{i}{}{\bm{T}}_{\text{obj}} = \prescript{i}{}{\bm{T}}_{\text{EE}, i} \cdot \prescript{\text{EE}, i}{}{\bm{T}}_\text{obj}.
\end{equation}
This relationship $\mathbf{x}_{\text{obj}} = \bm{h}_{\text{obj}}(\bm{q}_i)$ maps joint variables to the object pose $\mathbf{x}_{\text{obj}}  \in \mathbb{R}^7$, including position and quaternion-based orientation. The closed-chain constraint between the $i$-th and the $j$-th robot ensures that \( \bm{h}_{\text{obj}}(\bm{q}_{\mathrm{R}_i}) = \bm{h}_{\text{obj}}(\bm{q}_{\mathrm{R}_j}) \).

In this work, the dynamics of the manipulated objects are simplified, emphasizing coordinated motion planning of the multi-robot system rather than individual object dynamics. Specifically, the objects used in experiments are lightweight. However, incorporating a compensating torque term into the robots' control laws is straightforward, if dynamic properties of the objects and a desired load distribution are provided.

\subsection{Objectives}
\begin{figure}[htbp]
	\vspace{-2mm}
	\centering
	\includegraphics[width = 0.99\linewidth]{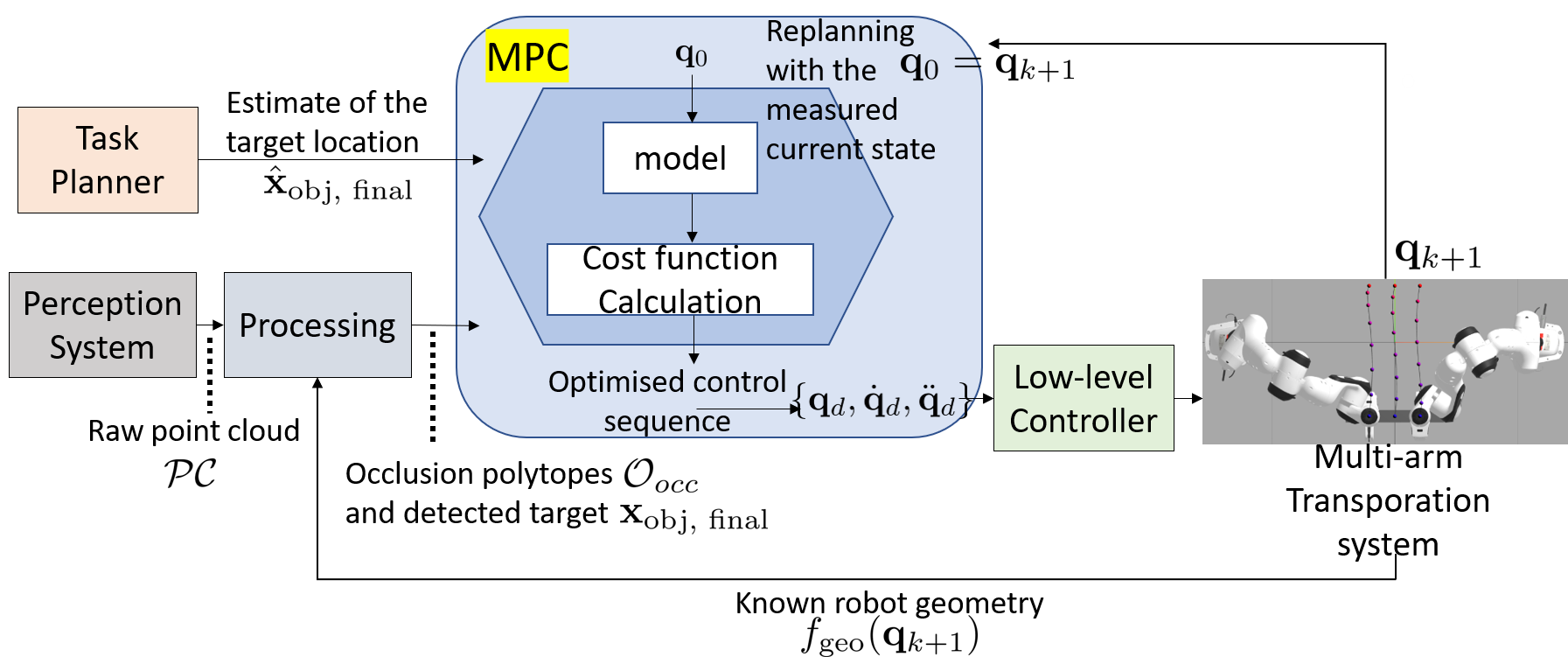}
	\vspace{-5mm}
	\caption{System overview with the MPC planner bridging task planning and control, generating motion references from real-time state and perception.}
	\label{fig:Control Diagram}
	\vspace{-2mm}
\end{figure}

The objective is to develop an MPC-based motion planner that enables the multi-robot system described above to navigate through cluttered environments. Positioned as an intermediate layer between the high-level task specifications and the low-level control layer, as shown in Fig~\ref{fig:Control Diagram}, the planner continuously updates its motion predictions and constraints in real time. It integrates the latest state information with perception data from vision systems to compute the motion references $\{\bm{q}_d, \dot{\bm{q}}_d, \ddot{\bm{q}}_d\}$ necessary for the robots' controllers.

The planner ensures the coordinated motion and whole-body obstacle avoidance, enabling the robots to handle both unmodelled static obstacles and dynamic obstacles. In scenarios with initially partial observations of the target and environment, the planner proactively guides the robot to positions that improve visibility and facilitate further perception. Overall, this MPC-based motion planner should adjust dynamically to real-time changes, enabling safe and efficient navigation for cooperative transportation.

\section{Proposed Motion Planner}
\subsection{Proposed Method}
We present a trajectory optimization approach to ensure a collision-free cooperative transportation in clutter for two 7-DoF FE cobots. We augment the robot joint states $\mathbf{q} = \begin{bmatrix} \bm{q}_{\mathrm{R}_1}^T, \bm{q}_{\mathrm{R}_2}^T \end{bmatrix}^T$ with the object pose $\mathbf{x}_{\text{obj}}$, forming the system state $\mathbf{z} = \begin{bmatrix} \bm{q}_{\mathrm{R}_1}^T, \bm{q}_{\mathrm{R}_2}^T, \mathbf{x}_{\text{obj}}^T \end{bmatrix}^T \in \mathbb{R}^{21}$. The components of $\mathbf{z}(t)$ are inherently coupled by the closed-chain constraint $\bm{h}_{\text{obj}}(\bm{q}_{\mathrm{R}_1}) = \bm{h}_{\text{obj}}(\bm{q}_{\mathrm{R}_2}) = \mathbf{x}_{\text{obj}}^T$.

The state trajectory $\mathbf{z}(t)$ is encoded using B-splines of degree $n$, parameterized by $M$ control points $\mathbf{c}_i$, as follows:
\begin{equation}
	\mathbf{z}(t) = \sum^{M-1}_{i = 0} \mathbf{c}_{i} B^{n}_{i}(t),
\end{equation}
where each control point $\mathbf{c}_i$ shares the same physical interpretation as the system state $\mathbf{z}$. The basis functions $B^{n}_{i}(t)$ are defined recursively using De Boor's algorithm \cite{wang2022continuous}.
Due to the inherent smoothness of B-splines, which are polynomials with $C^{n-1}$ continuity, the velocity and acceleration trajectories can be expressed as:
\begin{equation} \begin{aligned}&\mathbf{v}(t) = \sum^{M-2}_{i = 0} \mathbf{v}_i B^{n-1}_{v,i}(t), \ &\mathbf{a}(t) = \sum^{M-3}_{i = 0} \mathbf{a}_i B^{n-2}_{\mathbf{a},i}(t). \end{aligned} \end{equation}
High-order derivatives $B^{n-1}_{\mathbf{v},i}(t)$ and $B^{n-2}_{\mathbf{a},i}(t)$ of the original basis function $B^{n}_{i}$ are derived using the chain rule on the recursive formula. New control points $\mathbf{a}_i$ and $\mathbf{v}_i$ are computed from $\mathbf{c}_i$. Definitions and properties of B-splines are detailed in \cite{biagiotti2010b}. Following \cite{cai2024mpc}, we derive a time-independent trajectory representation using the normalized phase variable $s(t) = \displaystyle{\frac{t}{T}}$ for total duration \( T \).
\begin{equation*}
	\begin{aligned}
		&\mathbf{z}(s) = \sum^{M-1}_{i = 0} \mathbf{c}_{i} B^{n}_{i}(s), \\
		\mathbf{v}(s) = \sum^{M-2}_{i = 0} \frac{\mathbf{v}_i}{T} & B^{n-1}_{\mathbf{v},i}(s), \quad
		\mathbf{a}(s) = \sum^{M-3}_{i = 0} \frac{\mathbf{a}_i}{T^2} B^{n-2}_{\mathbf{a},i}(s).
	\end{aligned}
\end{equation*}
The OCP can be represented as a numerical optimization problem using the B-spline transcription as follows:

\begin{align}
	\underset{c_{0:M-1},\, T,\, \boldsymbol\varepsilon_{\text{fp}}}{\text{min}} \quad
	& \sum_{i=0}^{N} \left\| \frac{\mathbf{a}_i}{T^2} \right\|^2_{W_a}
	+ \left\| \mathbf{q}(s_i) - \frac{\mathbf{q}_{\min} + \mathbf{q}_{\max}}{2} \right\|^2_{W_m} \notag \\
	& + W_d T + W_{\text{vis}} \mathcal{C}_{\text{vis}} + \left\| \boldsymbol{\varepsilon}_{\text{fp}} \right\|^2_{W_{\text{fp}}} \notag \\
	\text{s.t.} \quad
	& \bm{h}_{\text{obj}}(\mathbf{q}(s_0)) = \mathbf{x}_{\text{obj, initial}} \notag \\
	& \bm{h}_{\text{obj}}(\mathbf{q}(s_N)) = \mathbf{x}_{\text{obj, final}} + \boldsymbol{\varepsilon}_{\text{fp}} \notag \\
	& \bm{h}_{\text{obj}}(\mathbf{q}(s_i)) = \mathbf{x}_{\text{obj}}(s_i) \notag \\
	& \mathbf{z}_{\min} \leq \mathbf{c}_i \leq \mathbf{z}_{\max} \label{eq:MPC_formulation_final}  \\
	& T \dot{\mathbf{z}}_{\min} \leq \mathbf{v}_i \leq T \dot{\mathbf{z}}_{\max} \notag \\
	& T^2 \ddot{\mathbf{z}}_{\min} \leq \mathbf{a}_i \leq T^2 \ddot{\mathbf{z}}_{\max} \notag\\
	& \textsc{score}(\mathbf{z}(s_k)) = 0. \notag
	\end{align}

Decision variables in the optimization framework include the trajectory duration $T$, control points $\bm{c}$, and slack variable $\boldsymbol\varepsilon_{\text{fp}}$. The B-spline representation of the trajectory is evaluated at N+1 collocation points, uniformly spaced in the equidistant knot vector $\{s_0, s_1, \dots, s_N \}$, representing the trajectory phases towards the goal. The key objectives guiding the motion planning process are summarized as follows:
\begin{itemize}
	\item \textbf{Trajectory Smoothness:} Smooth motion across the joint space and the manipulated object's task space is promoted by penalizing acceleration control points $\mathbf{a}_i$.
		
	\item \textbf{Agent Dexterity:} : Robotic agents are steered clear of joint limits to ensure dexterous manipulation, enforced by a cost term $\displaystyle{\Vert \mathbf{q}(s_i) - \displaystyle{\frac{\mathbf{q}_{\text{min}} + \mathbf{q}_{\text{max}}}{2}}\Vert}^2_{{W_m}}$, which minimizes deviations w.r.t. the joint range center.
	
	\item \textbf{Trajectory Duration:} Trajectory duration $T$ is explicitly considered to balance efficiency and task feasibility.

	\item \textbf{Visibility Maximization:} A vision cost $\mathcal{C}_\text{vis}$ is incorporated to ensure eye-in-hand cameras maintain informative views of the target, especially in the early stages when it is only partially visible due to clutter or occlusion. Details are provided in Section~\ref{ch:targetVis}.

	\item \textbf{Object Pose Constraints:} Equality constraints fix the initial and final poses of the object, with a tolerance term $\boldsymbol\varepsilon_{\text{fp}}$ allowing flexibility for uncertain targets. This setup facilitates exploration-driven motion that enhances target detection, complementing the vision cost $\mathcal{C}_\text{vis}$.
	
	\item \textbf{Closed-Chain Kinematic Constraint:}
	The rigid grasping condition is strictly enforced via: $\bm{h}_{\text{obj}}(\mathbf{q}(s_i)) = \mathbf{x}_{\text{obj}}(s_i)$, ensuring that the object remains constrained by the coupled motion of the multi-arm system.
	
	\item \textbf{Kinematic Limits:} Joint and workspace constraints are managed using inequalities with upper bounds $\mathbf{z}_{\text{max}} = [\mathbf{q}_{\text{max}}^T, \mathbf{x}_{\text{obj, max}}^T]^T$. $\mathbf{q}_{\text{max}}^T$ sets joint limits, while $\mathbf{x}_{\text{obj, max}}$ is set to infinity as the closed-chain constraint inherently limits the object's motion, enhancing computational efficiency without impairing the motion feasibility.
	
	\item \textbf{Whole-Body Collision Avoidance:} Collision avoidance is enforced as a hard constraint, ensuring zero proxy collision scores for all rigid body groups at collocation points, indicating collision-free configurations. Detailed methods and formulations are discussed in \ref{ch:ColliAvoid}.
\end{itemize}

\subsection{Point Cloud Processing}
\label{ch:PointCloud}
\begin{figure}
	\centering
	\vspace{-2mm}
	\includegraphics[width=0.97\linewidth]{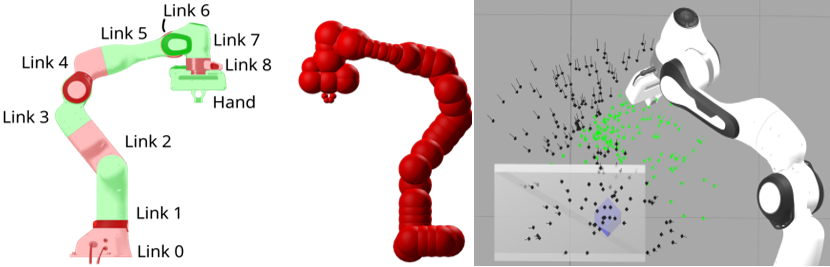}
	\caption{Left: FE cobot link indices and collision geometries~\cite{drake}. Right: A white obstacle polytope blocks both stationary and eye-in-hand cameras from viewing the blue placing spot. Green arrows show feasible eye-in-hand camera poses, computed via IK and selected through visibility reasoning in \ref{ch:targetVis} for clear target views. Black arrows: occluded or unreachable poses.}
	\label{fig:robot-link/colli-geometry_and_target-visibility-example}
	\vspace{-0.6cm}
\end{figure}
In our perception system, two cameras--a global camera and an eye-in-hand camera—capture point clouds for motion planning. These point clouds, representing real--time snapshots, pose challenges such as the lack of historical context and difficulty in tracking occluded objects. We address this by merging views from both cameras to enhance workspace representation and navigation in dynamic environments, filtering out irrelevant data using the robot's known geometry.

Our algorithm refines point cloud data by utilizing robot geometry modeled as spheres, shown in Fig.~\ref{fig:robot-link/colli-geometry_and_target-visibility-example}. It involves:
\begin{enumerate}
	\item \textbf{Point Cloud Filtering:} Calculating distances between each sphere and point cloud elements, retaining points that exceed predefined sphere radii to filter out parts representing the robot's own geometry.
	\item \textbf{Geometrical Constructs:} Using DBSCAN \cite{ester1996density} to segment filtered clouds into convex hulls for environmental delineation; occlusion polytopes are created by projecting convex hull vertices to outline visible areas.
	\item \textbf{Occlusion Comparison and Integration:} Combining current and historical visibility data from point clouds to maintain continuity in environment representation.
	\item \textbf{Dynamic Occlusion Detection:}  Each robot sphere is projected as a cone from camera perspectives to define occlusion zones. The cone's aperture angle \(\theta\) is calculated from the sphere's radius and distance from the camera, determining which points, previously visible, may now be occluded due to the robot's movement. 
\end{enumerate}

Further algorithmic details and pseudocode will be made available on Github upon acceptance (to preserve double-blind review integrity).




\subsection{Collision Avoidance}
\label{ch:ColliAvoid}

Motion planning in configuration space necessitates understanding both collision-free and in-collision regions. For highly redundant manipulators, the complexity of C-space topologies makes deriving explicit collision boundaries analytically challenging. We address this by adopting a data-driven approach using the kernel perceptron-based method, \textit{DiffCo} \cite{zhi2022diffco}, refined and extended to suit our specific needs.

\subsubsection*{DiffCo Method}
\textit{DiffCo} classifies configurations at the boundary between collision-free and in-collision states using a set of $M$ representative configurations \( \bm{S} \in \mathbb{R}^{M \times D} \). It computes collision scores for a query configuration \( \bm{q}_i \) as:
\begin{equation}
	\textsc{score}(\bm{q}_i) = k(\bm{q}_i, \bm{S})\mathbf{W}_{\bm{S}},
\end{equation}
where \( k(\bm{q}_i, \bm{S}) \) is a kernel function assessing similarity between \( \bm{q}_i \) and \( \bm{S} \). Training involves optimizing support configurations \( \bm{S} \) and weights \( \mathbf{W} \in \mathbb{R}^{M \times c} \) using a dataset \( (\bm{X}, \bm{Y}) \), where \( \bm{X} \) contains training configurations and \( \bm{Y} \) indicates the ground truth collision status. Kernel matrix \( \mathbf{K} \) and predicted scores (hypothesis matrix) \( \mathbf{H} \) are computed as:
\begin{equation}
	\mathbf{H} = \mathbf{K} \mathbf{W},
\end{equation}
minimizing discrepancies between \( \bm{Y} \) and \( \mathbf{H} \). The sparse update rule used in training focuses on minimizing non-zero support points to enhance the computational efficiency \cite{zhi2022diffco}. After training, configurations deemed redundant--where the corresponding weights in \( \mathbf{W} \) are zero--are pruned from the support set. This results in a compact, informative set that clearly delineates collision boundaries in the C-space.

In this work, we use the polyharmonic kernel, defined as:
\[
k_{\text{PH}}(x, x') \!= \!
\begin{cases} 
	r^k & \! \text{if } k \text{ is odd}, \\
	r^k \ln(r) & \! \text{if } k \text{ is even}, \, \text{where } r = \|x - x'\|_2.\\
\end{cases}
\]
where $x$ and $x'$ represent configuration vectors (not to be confused with the Cartesian pose variable $\mathbf{x}$ used earlier). Applied across $M$ forward kinematics control points, this kernel captures robot geometry in Cartesian space and provides a consistent, differentiable collision score, contrasting with the non-intuitive output distribution from Gaussian kernels \cite{das2020learning}. Hence, its gradients meaningfully indicate movement away from collisions in C-space. The kernel function for configuration similarity measurement is:
\begin{equation}
	k_{\textsc{FK}}(\bm{q}_i, \bm{q}_{\bm{S}})  = \frac{1}{M} \sum_{m=1}^{M} k_{\textsc{PH}} \left( \textsc{FK}_m(\bm{q}_i), \textsc{FK}_m(\bm{q}_{\bm{S}}) \right).
	\label{eq:org FK Kernel}
\end{equation}
To adapt to dynamic environmental changes and real-time perception updates affecting $C_\text{obs}$ and $C_\text{free}$, \textit{DiffCo} incorporates a sampling-based active learning strategy that avoids costly re-computation. It operates in two phases:

\begin{itemize} 
	\item \textbf{Exploitation}: Samples near existing support points with a Gaussian distribution to adapt to minor boundary shifts from slow-moving obstacles, refining support vectors to maintain accurate boundary representation.
	\item \textbf{Exploration}: 
	Detects new, unmodeled obstacles by random sampling across the joint space. New configurations are added to the support vectors, with zeros integrated into previous weights and hypothesis matrices to maintain model integrity without premature bias.
\end{itemize}
These updates effectively adapt the support vector set $\bm{S}$ to changing conditions without restarting the learning process from the scratch. More details are described in \cite{zhi2022diffco}.

\subsubsection*{Limitations}
Two primary limitations were identified in our application with \textit{DiffCo} method: ineffective overfitting management, where \textit{DiffCo} only excludes support points beyond a threshold, and the use of a forward kinematics kernel for all links simultaneously, which can obscure critical collision information by overwhelming it with irrelevant data.
\begin{figure}[htbp]
	\vspace{-6mm}
	\centering
	\includegraphics[width = 0.495 \textwidth]{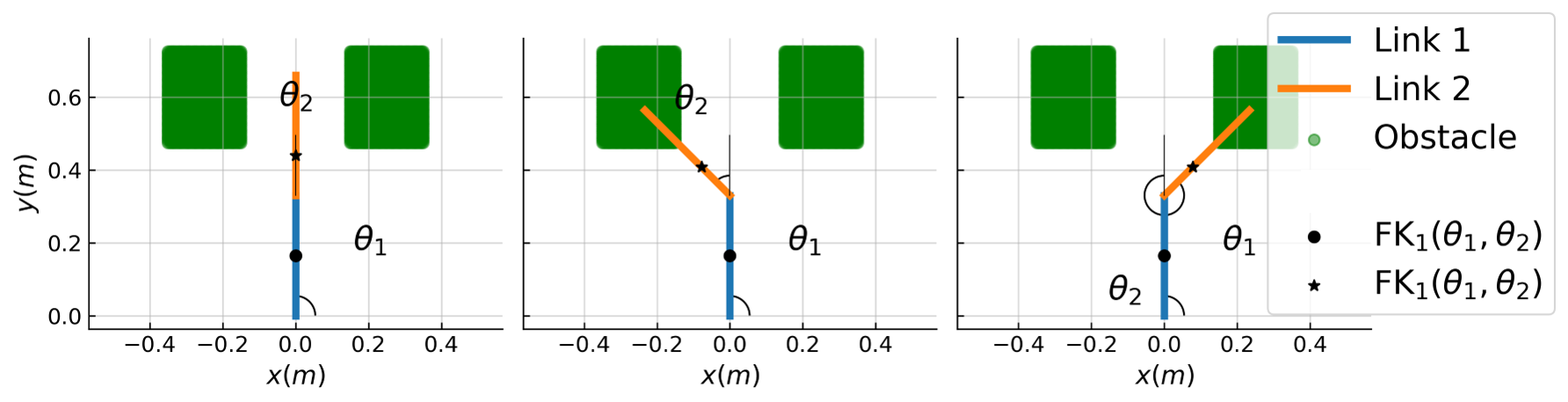}
	\vspace{-6mm}
	\caption{Example configurations of a two-link planar robot.}
	\label{fig:3 Configurations}
	\vspace{-3mm}
\end{figure}

Consider a two-link planar robot in three different configurations, shown in Fig.~\ref{fig:3 Configurations}. The kernel perceptron method employed with the forward kinematics kernel in (\ref{eq:org FK Kernel}), exhibits significant drawbacks. Control points, positioned centrally on each link, feed into the kernel, leading to:
\vspace{-1.0ex}
\[
\mathbf{K}_{\text{FK}} = \begin{bmatrix}
	1.0000 & 0.8269 & 0.8269 \\
	0.8269 & 1.0000 & 0.5567 \\
	0.8269 & 0.5567 & 1.0000
\end{bmatrix}.
\]
Averaging control points leads to high similarity scores, diminishing the kernel's discriminative ability. For example, high $\mathbf{K}_{\text{FK}}(1,2)$ and $\mathbf{K}_{\text{FK}}(1,3)$ values imply proximity between configurations despite varying collision relevance.

\begin{figure}[htbp]
	\centering
	\vspace{-2mm}
	\includegraphics[width=0.495\textwidth]{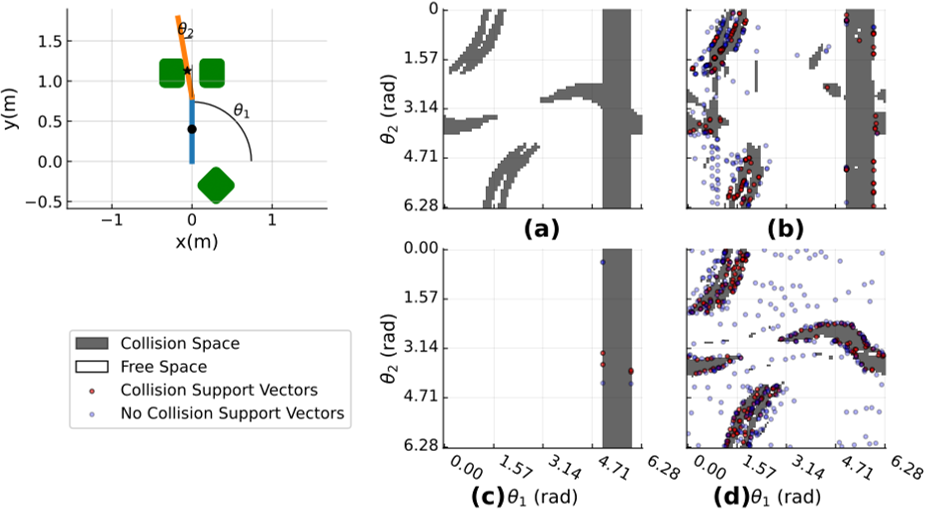}
	\vspace{-5mm}
	\caption{Comparison of original and modified algorithms in configuration space.
	Left: Robot configuration and obstacles in Cartesian space.
	(a) Ground truth configuration space (black: collision, white: free).
	(b) Original DiffCo learned space, showing excessive support vector clustering.
	(c)-(d) Modified algorithm learned spaces for joints 1 and 2, showing clearer boundaries and fewer support vectors.}
	\label{fig:Configuration Space DiffCo}
	\vspace{-4mm}
\end{figure}

Furthermore, \textit{DiffCo} handles overfitting by removing excess support points, which slows convergence and limits generalization, leading to imprecise collision predictions with false positives and negatives. As shown in Fig.~\ref{fig:Configuration Space DiffCo}(b), these issues deteriorate with new obstacles; the original DiffCo algorithm exhibits excessive clustering of support vectors around detailed collision zones without effective pruning, leading to misrepresentation of complex collision boundaries compared to the ground truth in Fig.~\ref{fig:Configuration Space DiffCo}(a). Adjusting kernel parameters can offer limited improvement but becomes impractical as the system complexity increases.


\subsubsection*{Algorithm Enhance}
We replaced the unified forward kinematics kernel with segmenting the robot into discrete collision groups, each with independently learned support vectors. Specifically, groups are defined as follows, as shown in Fig.~\ref{fig:robot-link/colli-geometry_and_target-visibility-example} :
\begin{itemize}
	\item Link 1 and 2
	\item Link 3 and 4
	\item Link 5
	\item An assembly comprising links 6, 7, 8, the gripper hand, fingers, and workpiece. 
\end{itemize}
Each group independently processes its collision constraints:
\begin{equation}
	\textsc{score}_{\text{g}_i}(\mathbf{x}(s_k)) = 0,
\end{equation}
A higher support vector limit is assigned to groups involved in dynamic activities, improving collision score accuracy in critical regions. This ensures a detailed boundary representation where needed, without significantly increasing computational load, leveraging parallel processing. Tests with FE cobots confirm faster processing and better generalization.

To mitigate overfitting, we refine the support vector set through iterative pruning based on the Gram matrix $\mathbf{K}_{\text{SV}}$, evaluated before each iteration to capture vector similarities. Vectors exceeding a threshold are removed, with thresholds adapted by collision status: lower for collision-free samples to reduce redundancy. This selective pruning enhances model efficiency and generalization to unseen environments.

Fig.~\ref{fig:Configuration Space DiffCo}(c) and (d) demonstrate the modified algorithm's enhanced accuracy and efficiency in the same planar robot case, requiring only 7 support vectors for the first link and 505 for the second--significantly fewer than the 563 vectors used by the original method. This improvement in computational efficiency becomes even more manifest based on our experiments with the FE cobots.

The initial active learning algorithm in Section~\ref{ch:ColliAvoid} often caused training weight ``explosions'', leading to numerical errors as weights neared float limits and updates stalled. Factors such as sample counts, noise, kernel parameters, and environment variations compounded these issues, making weight management cumbersome. Moreover, in multi-robot transport scenario, closed-chain constraints inherently restrict the feasible workspace, rendering broad sampling inefficient with high rejection rates and poor feasible-space coverage.

To address these challenges, we refine the active learning algorithm to reset weights and hypothesis matrices at each iteration, improving adaptability to dynamic changes. Support vectors are dynamically managed by pruning non-contributory ones to maintain model relevance and efficiency. The learning cycle is divided into an exploitation phase--refining decision boundaries--and an exploration phase with biased sampling based on past trajectories. This strategy optimizes sampling for cooperative tasks while reducing computational load. A detailed explanation and pseudocode will be released on Github upon acceptance.


\subsection{Target Visibility}
\label{ch:targetVis}

To enhance the target visibility when the placement spot is obscured, we employ a strategy adapted from our proxy collision detector, optimized for visibility. This involves a visibility score function, similar to the collision score, ensuring unobstructed line-of-sight (LoS) for the eye-in-hand camera to the target. Integrated into the MPC, this differentiable function dynamically adjusts trajectories for optimal visibility. The process begins with two-stage sampling to generate potential camera positions around the estimated spot, \( \bm{p}_{\text{place}} \). For each sampled position \( \bm{p}_i \), the algorithm:
\begin{itemize}
 \item Computes a potential camera orientation $\bm{R}_{\text{cam}}$ aligning directly with $\bm{p}_{\text{place}} $, forming the desired pose $\bm{T}_{\text{cam}}$.
 \item Validates these poses through: \\
 \quad - \textbf{Inverse Kinematics $\mathbf{q}_{\text{cam}} = \bm{h}^{-1}(\bm{T}_{\text{cam}})$:} Ensures the robot can physically reach the pose. \\
 \quad - \textbf{Visibility Check:} Confirms an unobstructed LoS with the obtained obstacle polytopes in Section \ref{ch:PointCloud}.
\end{itemize}
Mathematically, we define the visibility function as:
$$
\text{Visibility}(T_{\text{cam}}) = 
\begin{cases} 
	1 & \text{if IK feasible and no obstructions,} \\
	0 & \text{otherwise}.
\end{cases}
$$
Camera and wrist positions from valid \( \{\bm{T}_{\text{cam}}\} \) sets serve as control points for computing a forward kinematics kernel matrix \( \mathbf{K}_{\text{FK}} \), which prioritizes regions dense with feasible camera poses for optimization. These configurations, termed ``visibility support vectors'', are similar to methods used in proxy collision detector and guide the system toward the most advantageous poses for optimal target visibility. Fig.~\ref{fig:robot-link/colli-geometry_and_target-visibility-example} demonstrates this, displaying camera orientations and kinematic feasibility, with green arrows indicating unobstructed views. A detailed explanation and pseudocode will be released on Github upon acceptance.

\section{Results}
\subsection{Overview}
\begin{figure}[thbp]
	\centering
	\includegraphics[width=1\linewidth]{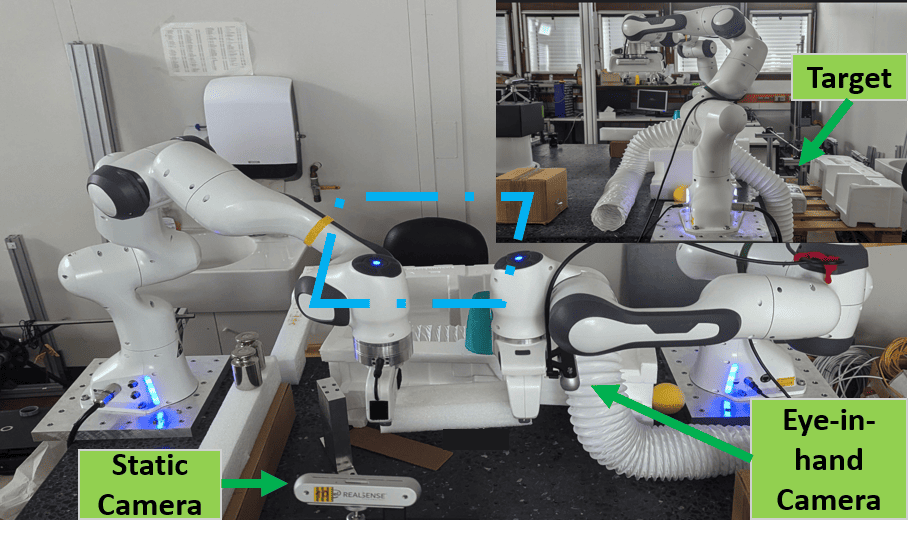}
	\vspace{-6mm}
	\caption{Experimental setup. Main: Cooperative multi-arm platform with the perception system. Highlighted area (partially occluded) marks the target region, where a wooden frame with an AprilTag is placed behind the obstacle. Inset: Side view clearly showing the placing target behind clutter.}
	\label{fig:exp setup}
	\vspace{-3mm}
\end{figure}

The proposed motion planning framework was evaluated through extensive simulations and real-world experiments, using a setup with two FE cobots and a perception system comprising a fixed and an eye-in-hand Intel RealSense D455 depth camera. The overall setup is shown in Fig.~\ref{fig:exp setup}. Simulations were conducted in Drake~\cite{drake}, and real-world trials in clutter validated the planner's ability to ensure dynamic feasibility, perform online re-planning, and explore unmodeled spaces. The experiments also assessed the real-time capability of the perception system and its integration with the motion planner, including point cloud processing to construct obstacle polytopes for collision avoidance.

Both manipulators were controlled via a single PC running Ubuntu 20.04 with the PREEMPT RT patch, equipped with an AMD Ryzen Threadripper 3960X CPU and an NVIDIA RTX A5000 GPU. The MPC planner and perception modules ran on this unit, communicating with the robots through ROS. The planner, operating on a fixed interval, continuously received robot states and sent back motion references executed by joint trajectory controllers. The framework was implemented and optimized in CasADi~\cite{Andersson2019}, leveraging code generation and a compiled IPOPT~\cite{nocedal2009adaptive} solver, with MA57 as the linear solver to improve computational efficiency.

Representative results under dynamic and cluttered conditions are presented below to demonstrate system performance and planner robustness. A supplementary video of the experiments is included with the submission.

\vspace{-2mm}
\subsection{Scenario I}
\vspace{-0.5mm}
In this scenario, the system must handle a placing spot whose pose is initially estimated from partial observations. A series of experiments validate the motion planner's ability to trigger exploratory movements to accurately locate the target, while maintaining collision avoidance in a static environment perceived through continuous real-time sensory updates.

\begin{figure}[thbp]
	\vspace{-2mm}
	\centering
	\includegraphics[width=0.95\linewidth]{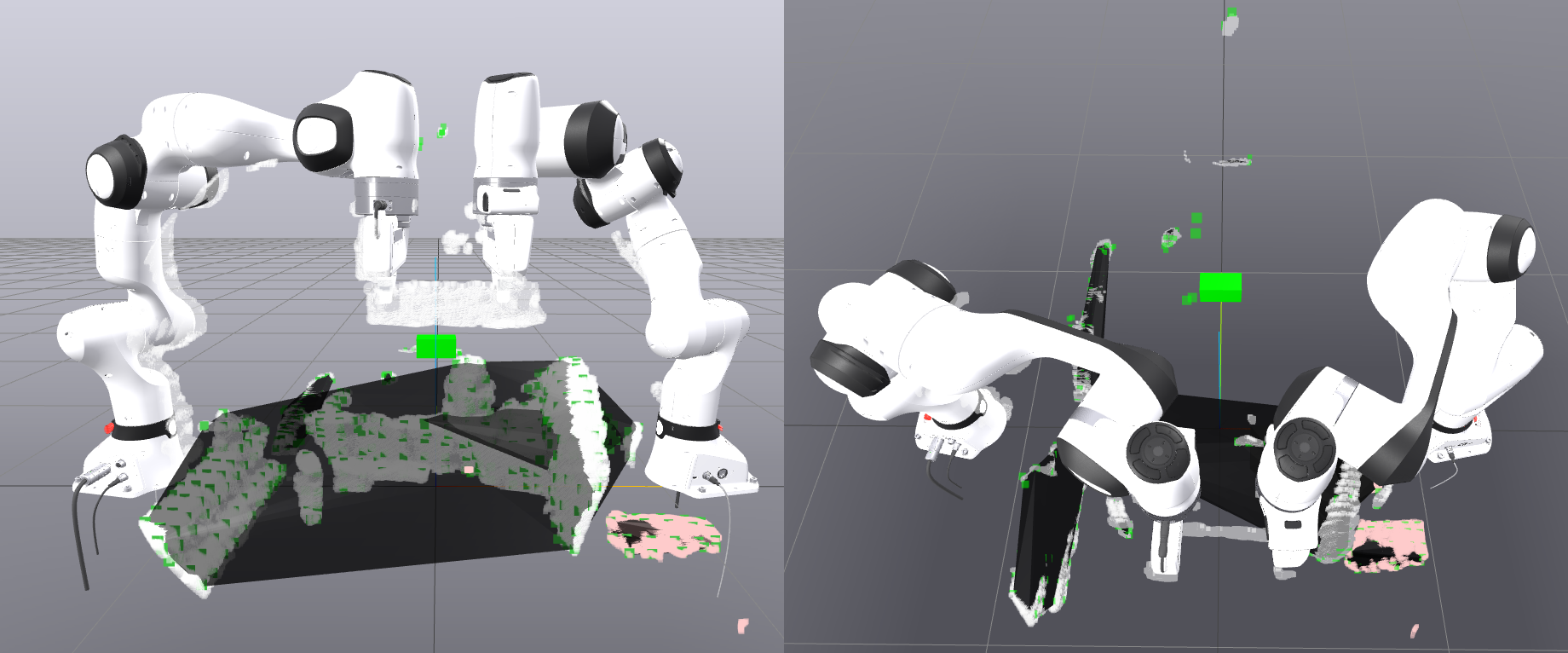}
	\caption{ 
		Environment configuration and perceived workspace at trajectory start. White and pink point clouds originate from the static and eye-in-hand cameras, respectively, after geometric filtering (\ref{ch:PointCloud}), as visible around the obstacle and robot bodies. Green dots represent processed points used to construct the black translucent occlusion polytopes integrated into the motion planning constraints in (\ref{eq:MPC_formulation_final}). Right: Top view showing the limited initial visibility near the target due to occlusions. This motivates subsequent exploration and online updating capabilities.}
	\label{fig:scenarioI_camera_perspective}
\end{figure}

The scenario begins with the cooperative robots initiating movement, as shown in Fig.~\ref{fig:scenarioI_camera_perspective}, displaying the initial point cloud. The initial estimate of the placing spot, marked with an AprilTag on a wooden frame, is indicated by a green box.

\begin{figure}[htbp]
	\centering
	\vspace{-2mm}
	\includegraphics[width=0.875\linewidth]{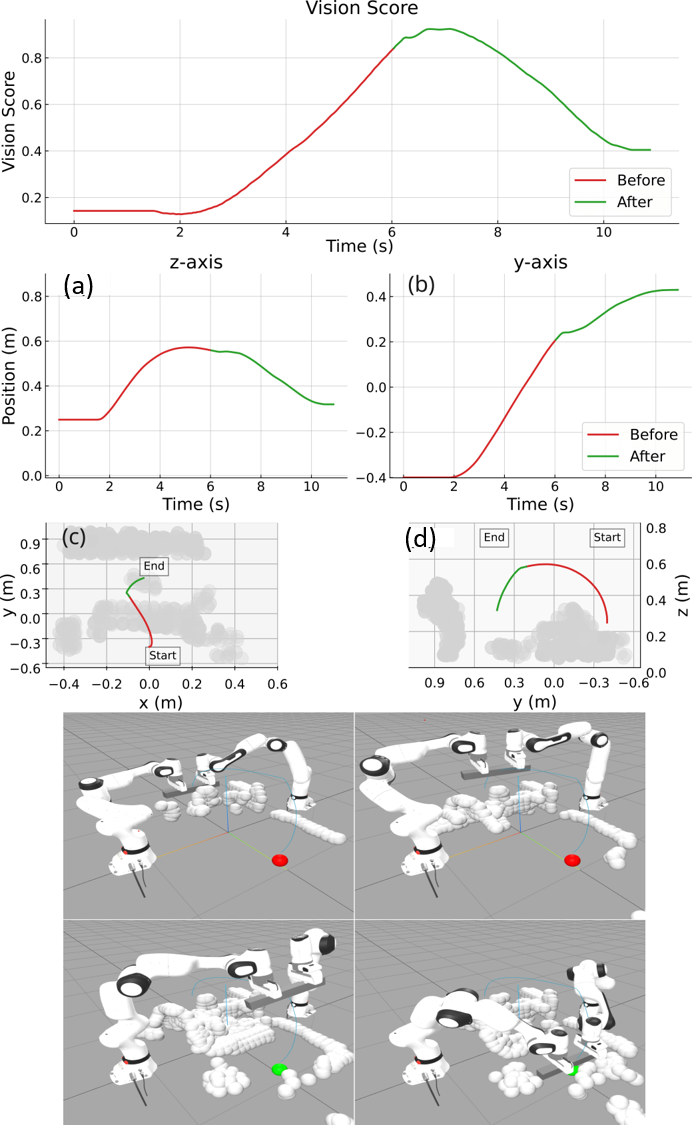}
	\vspace{-1mm}
	\caption{Experimental results from Scenario I. (Top) Visibility score over time. (a)-(b) Workpiece positions along z- and y-axes, color-coded by visibility. (c)-(d) Side views of executed Cartesian trajectories (xy and yz--planes). (Bottom) Experimental snapshots visualized in Drake. White spheres represent obstacle polytopes (see Section~\ref{ch:PointCloud}), convexified for efficient obstacle avoidance. Their changes across snapshots reflect continuous real-time updates from the dual-camera perception system.}
	\label{fig:scenarioI_results}
	\vspace{-6mm}
\end{figure}

Fig.~\ref{fig:scenarioI_results} presents results from a representative experiment. The top plot shows the visibility score defined in (\ref{eq:MPC_formulation_final}), computed as the similarity between the current eye-in-hand camera pose and the “visibility support vectors” in \ref{ch:targetVis}. The plot is color-coded: red for partial visibility, green once sufficient visibility is achieved—demonstrating the exploratory behavior induced by our MPC planner. The bottom snapshots are key frames captured from experimental data rendered in Drake, depicting the transition of the target marker from red (partially visible) to green (fully visible).

The eye-in-hand camera configuration's real-time capabilities provide continuous environmental updates, crucial for dynamically navigating and adapting to spatial constraints. By comparing Fig.~\ref{fig:scenarioI_camera_perspective} and the bottom snapshots of Fig.~\ref{fig:scenarioI_results}, the dual-camera perception system's effectiveness becomes evident: initial visibility around the placing target is limited by occlusions, but exploration dynamically updates the environment model, enabling safe planning in cluttered, partially unknown workspaces. Fig.~\ref{fig:scenarioI_results}(a)-(d) illustrates the trajectories executed by the robots while maneuvering the jointly held workpiece. Consistent with earlier plots, these trajectories use the same visibility-based color coding. Subfigures (a) and (b) depict the final trajectory's position changes along the y and z axes over time. Subfigures (c) and (d) represent side views (xy and yz planes) of the executed Cartesian trajectory segment. Notably, the trajectory exhibits slight  ``wobbles'' near completion--resulting from the MPC planner's finer adjustments as it navigates closely spaced obstacles in a cluttered environment. This behavior underscores the importance of the terminal constraint in our MPC formulation during critical operational phases. Despite local adjustments, the trajectory maintains overall smoothness due to the B-spline transcription employed, balancing the smoothness and responsiveness to real-time environmental variations.
\begin{figure}[bhtp]
	\centering
	\includegraphics[width=0.95\linewidth]{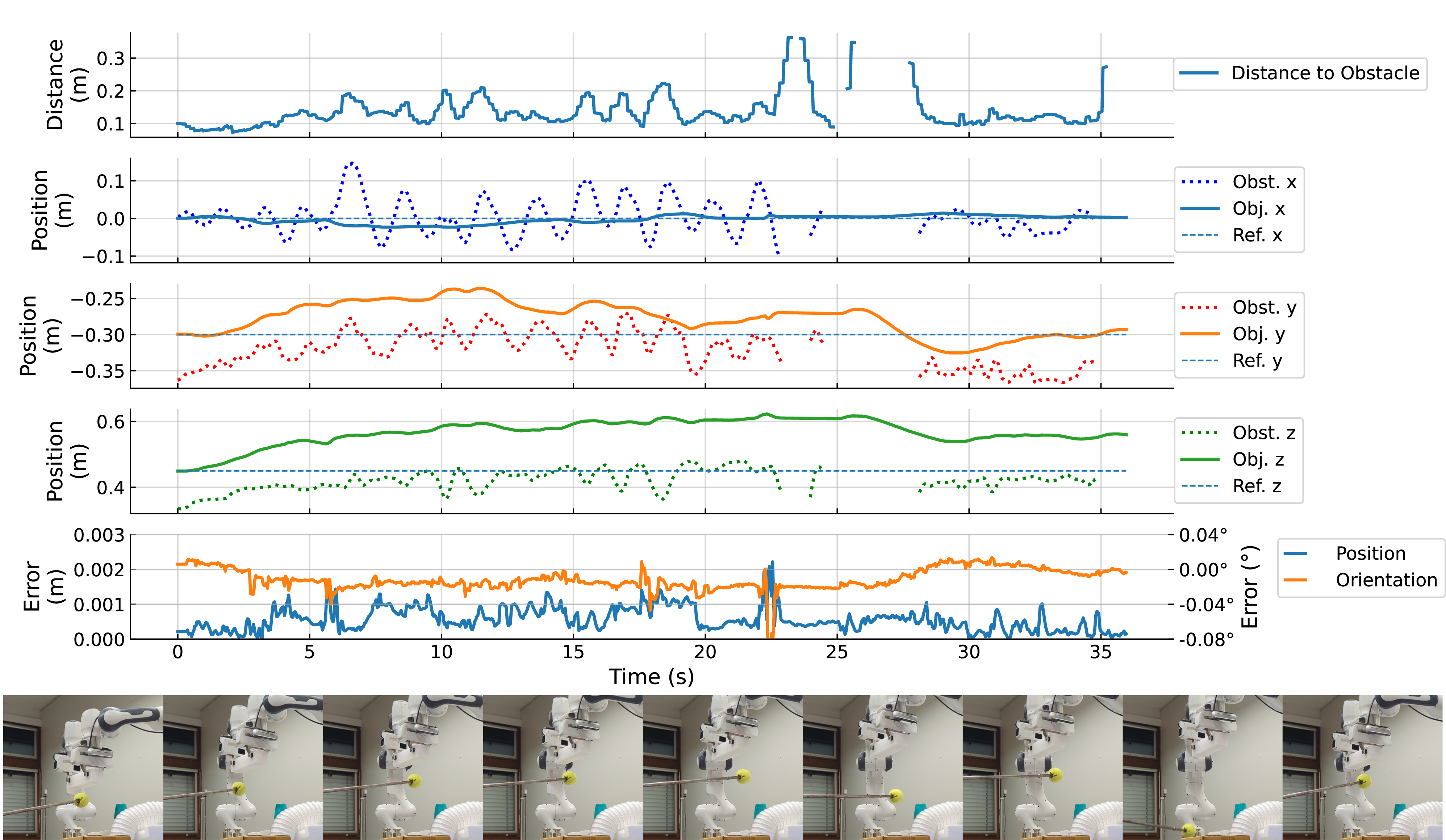}
	\vspace{-3mm}
	\caption{Experimental results of Scenario II. Top: quantitative performance plots. Bottom: key visual snapshots of the task execution. See supplementary video for full execution.}
	\label{fig:scenarioII_results}
	\vspace{-4mm}
\end{figure}

\vspace{-1mm}
\subsection{Scenario II}
\label{Scenario II}
\vspace{-0.5mm}
In this scenario, the motion planner is evaluated for its ability to ensure the interaction safety when a human enters the shared workspace. A tennis ball simulates human intrusion, triggering a safety response governed by MPC constraints--demonstrating adaptive avoidance without compromising task coordination, similar to the setup in \cite{he2022distributed}. While this experiment uses a proxy object, full-body human motion can also be integrated via detection systems, see \cite{cai2023probabilistic}.

As a human enters the workspace, the planner switches to a temporary "replanning mode," shifting focus from task execution to interaction safety. In this mode, the MPC's start and end pose constraints are both reset to the object's current pose, preserving the last known safe configuration. The end constraint is softened using slack variables $\boldsymbol\varepsilon_{\text{fp}}$ in (\ref{eq:MPC_formulation_final}), enabling the planner to adaptively generate evasive trajectories in response to the intrusion. This strategy maintains operational readiness in cluttered, unmodeled environments, as the pre-intrusion pose is known to be safe. Once the human exits, the system can seamlessly resume its original task without compromising the coordination or dexterity.

Experimental results, depicted in Fig.~\ref{fig:scenarioII_results}, illustrate the system's response to dynamic human movements. The bottom row provides key frames highlighting significant interaction moments. The first row shows the object's distance to obstacles over time; discontinuities occur when the obstacle moves outside the perception system's view. Rows two to four show object trajectory adaptations along the $x$, $y$, $z$ axes; disparities from reference poses at intrusion moments indicate adaptation enabled by slack variables $\boldsymbol\varepsilon_{\text{fp}}$ in (\ref{eq:MPC_formulation_final}). The second-to-last row shows minimal disparity between object poses independently estimated by each robot, confirming effective coordination and fulfillment of the closed-chain constraint. The consistency further suggests low conflicting internal force, even without detailed force analysis.

\section{CONCLUSIONS}
This letter presents a novel coarse-to-fine motion planning framework for multi-arm object transportation in cluttered, unmodeled environments. Centered on a shrinking-horizon MPC with B-spline transcription, the approach incrementally refines trajectories using real-time updates from a dual-camera system. Future work will extend this framework toward richer human-robot collaboration in dynamic, shared spaces, grounded in a deeper understanding of human presence, intent, and holistic cooperative behavior.

\addtolength{\textheight}{-12cm}   





\bibliographystyle{ieeetr}
\bibliography{references}


\end{document}